%% file: dynagan.tex
\documentclass[sigconf]{acmart}
\acmSubmissionID{487}

\usepackage{booktabs} 
\usepackage{graphicx}       
\usepackage{amsmath}        
\usepackage{multicol}       
\usepackage{multirow}       
\usepackage{color}          
\usepackage{colortbl}       
\usepackage{dsfont}         

\usepackage[ruled]{algorithm2e} 

\SetAlFnt{\small}
\SetAlCapFnt{\small}
\SetAlCapNameFnt{\small}
\SetAlCapHSkip{0pt}

\copyrightyear{2022}
\acmYear{2022}
\setcopyright{acmcopyright}\acmConference[SA '22 Conference Papers]{SIGGRAPH Asia 2022 Conference Papers}{December 6--9, 2022}{Daegu, Republic of Korea}
\acmBooktitle{SIGGRAPH Asia 2022 Conference Papers (SA '22 Conference Papers), December 6--9, 2022, Daegu, Republic of Korea}
\acmPrice{15.00}
\acmDOI{10.1145/3550469.3555416}
\acmISBN{978-1-4503-9470-3/22/12}

\begin{document}
\title[DynaGAN: Dynamic Few-shot Adaptation of GANs to Multiple Domains]{DynaGAN: Dynamic Few-shot Adaptation of GANs \\ to Multiple Domains}

\author{Seongtae Kim}
\orcid{0000-0002-5325-822X}
\affiliation{%
  \institution{POSTECH}
  \country{South Korea}
}
\email{seongtae0205@postech.ac.kr}

\author{Kyoungkook Kang}
\orcid{0000-0002-8964-1220}
\affiliation{%
  \institution{POSTECH}
  \country{South Korea}
}
\email{kkang831@postech.ac.kr}

\author{Geonung Kim}
\orcid{0000-0003-0806-6963}
\affiliation{%
  \institution{POSTECH}
  \country{South Korea}
}
\email{k2woong92@postech.ac.kr}

\author{Seung-Hwan Baek}
\orcid{0000-0002-2784-4241}
\affiliation{%
  \institution{POSTECH}
  \country{South Korea}
}
\email{shwbaek@postech.ac.kr}

\author{Sunghyun Cho}
\orcid{0000-0001-7627-3513}
\affiliation{%
  \institution{POSTECH}
  \country{South Korea}
}
\affiliation{%
  \institution{Pebblous}
  \country{South Korea}
}
\email{s.cho@postech.ac.kr}

\renewcommand{\shortauthors}{Kim et al.}

\input{macro}
\input{a_abstract}

%
%


%
%


\maketitle

\input{b_introduction}

\input{c_related_work}
\input{d_method}

\input{e_experiments}

\input{f_conclusion}

\bibliographystyle{ACM-Reference-Format}
\bibliography{egbib}

\end{document}

%% file: macro.tex
\def\MethodName{DynaGAN} 

\newcommand{\Eq}[1]  {Eq.\ (#1)}
\newcommand{\Eqs}[1] {Eqs.\ (#1)}
\newcommand{\Fig}[1] {Fig.\ #1}
\newcommand{\Figs}[1]{Figs.\ #1}
\newcommand{\Tbl}[1]  {Tab.\ #1}
\newcommand{\Tbls}[1] {Tabs.\ #1}
\newcommand{\Sec}[1] {Sec.\ #1}
\newcommand{\SSec}[1] {Sec.\ #1}
\newcommand{\Secs}[1] {Secs.\ #1}
\newcommand{\Alg}[1] {Alg.\ #1}
\newcommand{\etal}   {{\textit{et al.}}}

\newcommand{\setone}[1] {\left\{ #1 \right\}} 
\newcommand{\settwo}[2] {\left\{ #1 \mid #2 \right\}} 

\newcommand{\todo}[1]{{\textcolor{red}{#1}}}
\newcommand{\kkw}[1]{{\textcolor{cyan}{[KGU: #1]}}}
\newcommand{\kst}[1]{{\textcolor{magenta}{[kst: #1]}}}
\newcommand{\kstc}[1]{{\textcolor{magenta}{[kst(c): #1]}}}

\newcommand{\jy}[1]{{\textbf{\textcolor{MidnightBlue}{[JY] }}\textcolor{MidnightBlue}{#1}}}
\newcommand{\sean}[1]{{\textcolor{green}{sean: #1}}}
\newcommand{\sunghyun}[1]{{\textcolor[rgb]{0.6,0.0,0.6}{[sunghyun: #1]}}}
\newcommand{\sh}[1]{{\textcolor[rgb]{0.2,0.6,0.6}{[BAEK: #1]}}}
\newcommand{\kkang}[1]{{\textcolor[rgb]{0.6,0.0,0.1}{[kkang: #1]}}}
\newcommand{\change}[1]{{\color{red}#1}}
\newcommand{\origin}[1]{{\color{blue}#1}}
\newcommand{\bb}[1]{\textbf{\textit{#1}}}

\renewcommand{\topfraction}{0.95}
\setcounter{bottomnumber}{1}
\renewcommand{\bottomfraction}{0.95}
\setcounter{totalnumber}{3}
\renewcommand{\textfraction}{0.05}
\renewcommand{\floatpagefraction}{0.95}
\setcounter{dbltopnumber}{2}
\renewcommand{\dbltopfraction}{0.95}
\renewcommand{\dblfloatpagefraction}{0.95}


\newcommand{\Net}[1]{#1}
\newcommand{\Loss}[1]{$\mathcal{L}_{#1}$}
\newcommand{\cm}{\checkmark}
\newcommand\oast{\stackMath\mathbin{\stackinset{c}{0ex}{c}{0ex}{\ast}{\bigcirc}}}




%% file: a_abstract.tex
\begin{abstract}
Few-shot domain adaptation to multiple domains aims to learn a complex image distribution across multiple domains from a few training images.
A na\"ive solution here is to train a separate model for each domain using few-shot domain adaptation methods.
Unfortunately, this approach mandates linearly-scaled computational resources both in memory and computation time and, more importantly, such separate models cannot exploit the shared knowledge between target domains.
In this paper, we propose \MethodName{}, a novel few-shot domain-adaptation method for multiple target domains.
\MethodName{} has an adaptation module, which is a hyper-network that dynamically adapts a pretrained GAN model into the multiple target domains.
Hence, we can fully exploit the shared knowledge across target domains and avoid the linearly-scaled computational requirements.
As it is still computationally challenging to adapt a large-size GAN model, we design our adaptation module to be lightweight using the rank-1 tensor decomposition. 
Lastly, we propose a contrastive-adaptation loss suitable for multi-domain few-shot adaptation.
We validate the effectiveness of our method through extensive qualitative and quantitative evaluations.
\end{abstract}

\begin{teaserfigure}
   \centering
   \includegraphics[width=0.97\textwidth]{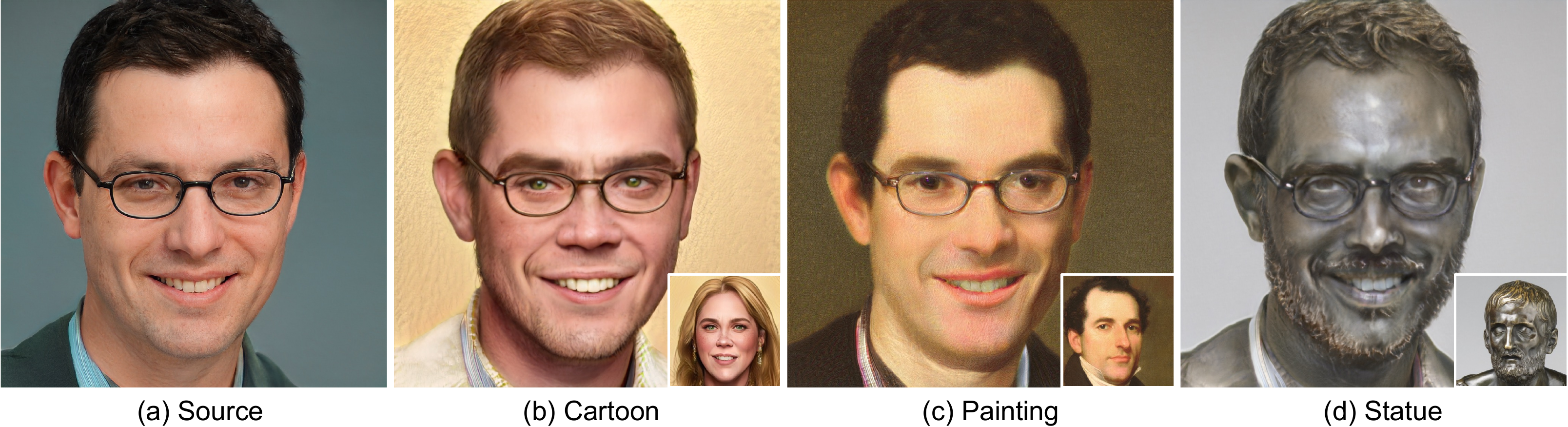}
   \vspace{-3mm}
   \caption[teaser]{
      We present \MethodName{}, a method that dynamically adapts a generator to each target domain for synthesizing multi-domain images using a single generator.
  (a) For a source image and ten target images in different domains (three of them shown in insets), we demonstrate successful domain-adaptation results: (b) cartoon, (c) painting, and (d) statue. 
Inputs in (c) and (d): The Metropolitan Museum of Art [Public Domain].
   }
   \label{fig:teaser}
\end{teaserfigure}

%% file: b_introduction.tex

\section{Introduction}
\label{sec:introduction}




Recent progress of generative adversarial networks (GANs) has opened up a new chapter in image synthesis~\cite{GANs,StyleGAN,StyleGAN2,StyleGAN3,BigGAN}, extending its application to various tasks including data augmentation~\cite{GANDataAug}, image restoration~\cite{GPEN,GLEAN,GFPGAN}, and image and video manipulation~\cite{IDinvert,BDinvert,psp,e4e,InterFaceGAN,Sefa,StyleCLIP,stitch}.
Unfortunately, this success has been mainly demonstrated on a few large-scale datasets such as human portraits~\cite{StyleGAN,CelebA}, because of the fundamental requirement of GANs on many training samples.
Thus, it is still challenging to apply GANs to diverse domains where data is scarce. 
While recent methods~\cite{StyleGAN-ada,DiffAug,DeceiveD} have alleviated the burden of data-intense GAN training, they still require hundreds of training images. 

To circumvent collecting a large-scale dataset, domain adaptation methods have been proposed~\cite{TGAN, MineGAN, BSA}.
Their core idea is to exploit a GAN model pretrained on a data-sufficient domain and finetune it onto a data-scarce domain where common knowledge exists between the two domains. 
Recent few-shot domain adaptation methods have demonstrated impressive success with less than ten training images~\cite{FS-ada, StyleGAN-nada, MTG}.
However, they assume that training images have similar appearances, 
and fail to handle training images with diverse domains as shown in \Fig~\ref{fig:qual_cmp}, generating an \emph{average style} of multiple domains.
One can bypass such failure by learning a separate model for each target domain, resulting in many models depending on the number of target domains. 
Unfortunately, this approach mandates linearly-scaled computational resources both in memory and computation time. 
More importantly, such separate models cannot exploit the shared knowledge between target domains, e.g. common structural composition in oil-painting portraits by different artists.
This results in the limited capability of domain-adapted models.

This paper proposes \MethodName{}, a novel few-shot domain adaptation method to handle multiple target domains. 
We depart from the separate-model principle and use a single GAN model pretrained on a data-sufficient domain.
To this end, we propose an adaptation module, a \emph{hyper-network} that dynamically adapts a GAN model to each target domain. 
This change allows us to avoid the problems of computational resources and overlooked shared knowledge between target domains.
Specifically, our adaptation module takes a target-domain condition vector as an input and modulates the weights of a GAN model, which acts as a dynamic domain adaptation to a target domain.
We carefully design our adaptation module to be light-weight and memory-efficient with rank-1 tensor decomposition. 
We train the adaptation module with a training dataset consisting of multiple target domains, each of which may have extremely few-shot images, e.g., only one image.
As we aim to maintain the unique style of each target domain, we further introduce a contrastive-adaptation loss to preserve the distinctive attributes of different target domains.
As shown in \Fig{\ref{fig:teaser}}, \MethodName{} allows us to synthesize images of diverse target domains, e.g., target domains of different drawing styles, and different animal species.

In addition to the reduced computational cost and diverse target domain adaptation, our method enjoys additional benefits. 
First, our approach performs domain adaptation using an adaptation module that modulates only the weights of the convolutional layers of a GAN model while preserving the GAN model itself intact.
As a result, the semantic editability of the original GAN model is naturally maintained in \MethodName{}, where the degree of adaptation can be easily controlled by scaling the amount of modulation from the adaptation module.
Second, as our adaptation module is trained on multi-domain images at once, it can effectively learn the shared knowledge across different domains, hence avoiding overfitting. 
Third, our approach allows interpolation of target-domain condition vectors, meaning that we can synthesize an image in a new target domain.
Our extensive experiments show that \MethodName{} achieves superior performance compared to the existing methods.

%% file: c_related_work.tex
\section{Related Work}




\paragraph{Generative Adversarial Networks.} 
Since GANs were firstly introduced by Goodfellow~\textit{et al.}~\shortcite{GANs}, synthesizing realistic images has been one of the core tasks and test-beds for GAN models. 
While we have witnessed the remarkable success of recent GAN-based synthesis methods including StyleGAN variants~\cite{StyleGAN,StyleGAN2,StyleGAN3}, one pitfall is that training a GAN model mandates a large-scale dataset, e.g., human portraits in FFHQ~\cite{StyleGAN} and church and car images in LSUN~\cite{LSUN}. 
Data augmentation is a promising way for overcoming the large-scale data problem by effectively increasing the training samples based on certain augmentation rules~\cite{StyleGAN-ada,DiffAug,DeceiveD}.
Unfortunately, state-of-the-art data-augmentation methods still require at least hundreds of images.


\paragraph{Domain Adaptation for GANs}
Domain adaptation is another way of detouring the large-scale data requirement of GANs~\cite{TGAN,MineGAN,UI2I}.
Once a GAN model is pretrained on a data-sufficient domain, domain-adaptation methods finetune the model on a data-scarce domain.
One of the major challenges of domain adaptation for GANs is that a GAN model easily overfits a few samples in the data-scarce domain.
This results in severe mode-collapse artifacts, meaning that an adapted GAN model can only synthesize virtually identical images~\cite{TGAN}.
Thus, few-shot domain adaptation of GANs has remained a challenging problem.
Recent methods have tackled this by means of target-weighted GAN latent space~\cite{MineGAN}, adapting batch statistics~\cite{BSA}, and regularizing the change of weights in a GAN model~\cite{EWC}.
Ojha~\textit{et al.}~\shortcite{FS-ada} propose to maintain the diversity of domain-adapted images as in the original GAN model by imposing a cross-domain consistency loss.
StyleGAN-nada~\cite{StyleGAN-nada} exploits the recently-proposed CLIP representation~\cite{CLIP}, which is a learned prior of natural images and text, in order to linearly translate the latent space from a source to a target domain maintaining the source diversity.
Mind-the-gap (MTG)~\cite{MTG} takes a step forward to achieve one-shot domain adaptation by preserving the source-domain diversity with reference-based CLIP regularizations.
Unfortunately, the existing domain-adaptation methods fail to handle a multi-domain few-shot dataset, for which our method is designed. 

\begin{figure*}[t]
\includegraphics[width=0.9\linewidth]{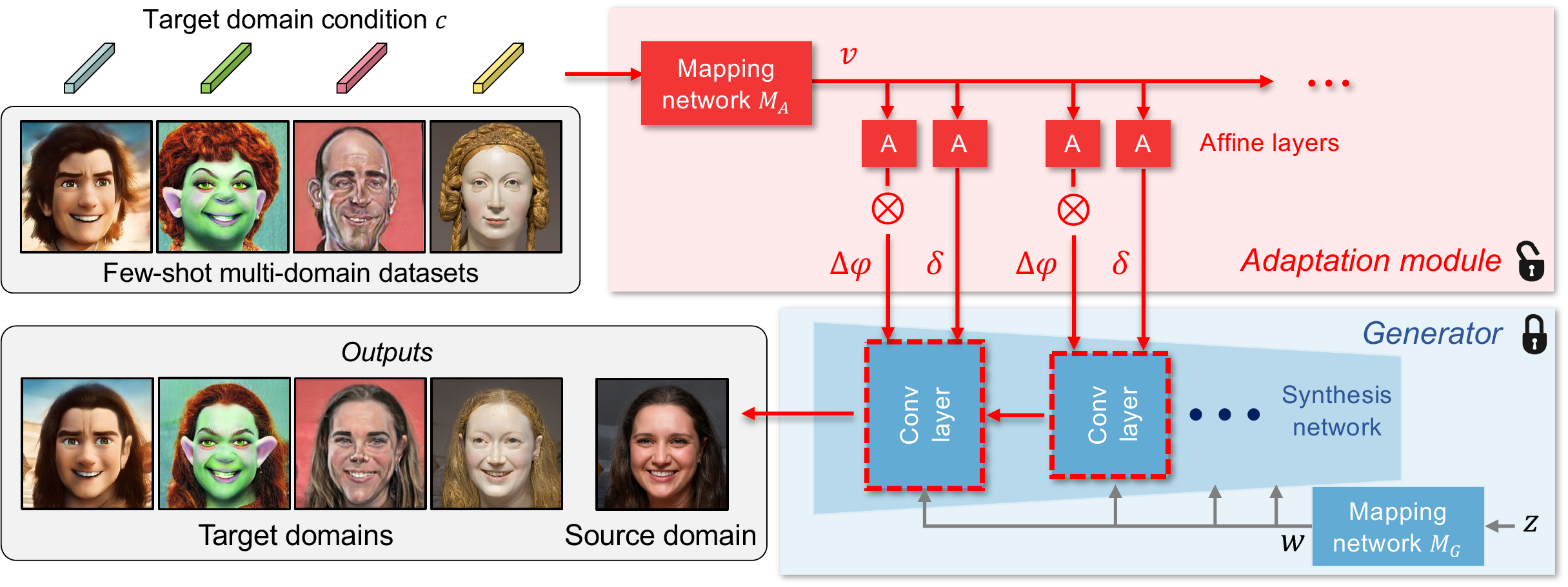}
\vspace{-5mm}
\caption{
\MethodName{} network consists of a generator pretrained on a source domain and an adaptation module that dynamically modulates the generator parameters. \MethodName{} takes a one-hot vector $c$ encoding a target domain as an input, projects it to a continuous representation $\upsilon$ through a mapping network $M_A$, and estimates modulation parameters $\Delta \varphi$ and $\delta$ through affine layers. These parameters modulate the weight of each convolutional layer in the generator. As a result, the generator is adapted to the target domain.
4th image in the few-shot multi-domain datasets: The Metropolitan Museum of Art [Public Domain].
}
\label{fig:framework}
\end{figure*}

\paragraph{Hyper-networks}
Hyper-networks, firstly proposed by Ha and Le~\shortcite{HyperNetwork}, are neural networks that modulate the parameters of other neural networks. 
Offering generalization and flexibility to existing neural networks, hyper-networks are being actively explored in many applications including neural architecture search~\cite{NASNet,HyperGAN}, multi-task learning~\cite{HyperFormer}, continual learning~\cite{HyperContinual}, semantic segmentation~\cite{HyperSeg}, and 3D modeling~\cite{DeepShape}. 
Most relevant to us, Alaluf~\textit{et al.}~\shortcite{HyperStyle} and Dinh~\textit{et al.}~\shortcite{HyperInverter} demonstrate that using a hyper-network to control a GAN model enables an out-of-representation input image, which cannot be accurately represented by a GAN model, to be successfully mapped to the valid representation space of the GAN. 
In this paper, we exploit a hyper-network for the few-shot multi-domain adaptation problem. 

%% file: d_method.tex
\section{\MethodName{}}






We aim to adapt a GAN model pretrained on a source domain to multiple target domains with only a few target images,
e.g., from the real-portrait FFHQ dataset~\cite{StyleGAN} to cartoon, anime, and painting with only one or two images for each.
To this end, we set three design goals for our method.
First, our model should be capable of synthesizing diverse images reflecting distinct target domain characteristics, not an average style of them.
Second, our model should be compact in model size in order to provide benefits over using a separate model for each target domain.
Lastly, our model should be controllable in choosing which target domain to use for image synthesis.
In the following, we present our framework \MethodName{} and its training strategy to meet these requirements.

\subsection{Framework}
\label{sec:framework}
\MethodName{} consists of a generator network and an adaptation module as shown in \Fig{\ref{fig:framework}}.
For the generator network, we adopt a StyleGAN2 generator~\cite{StyleGAN2} pretrained on a source domain.
The generator synthesizes an image from a GAN latent vector $z$, and the style of the synthesized image is controlled by a target domain condition $c$ through the adaptation module.
Here, the adaptation module acts as a hyper-network that dynamically modulates the weights of the generator according to a given target domain condition $c$.
As we achieve domain adaptation using the adaptation module while keeping the generator intact, the generator can still synthesize images of the original source domain by turning off modulation from the adaptation module.

\begin{figure}[t]
\includegraphics[width=0.95\linewidth]{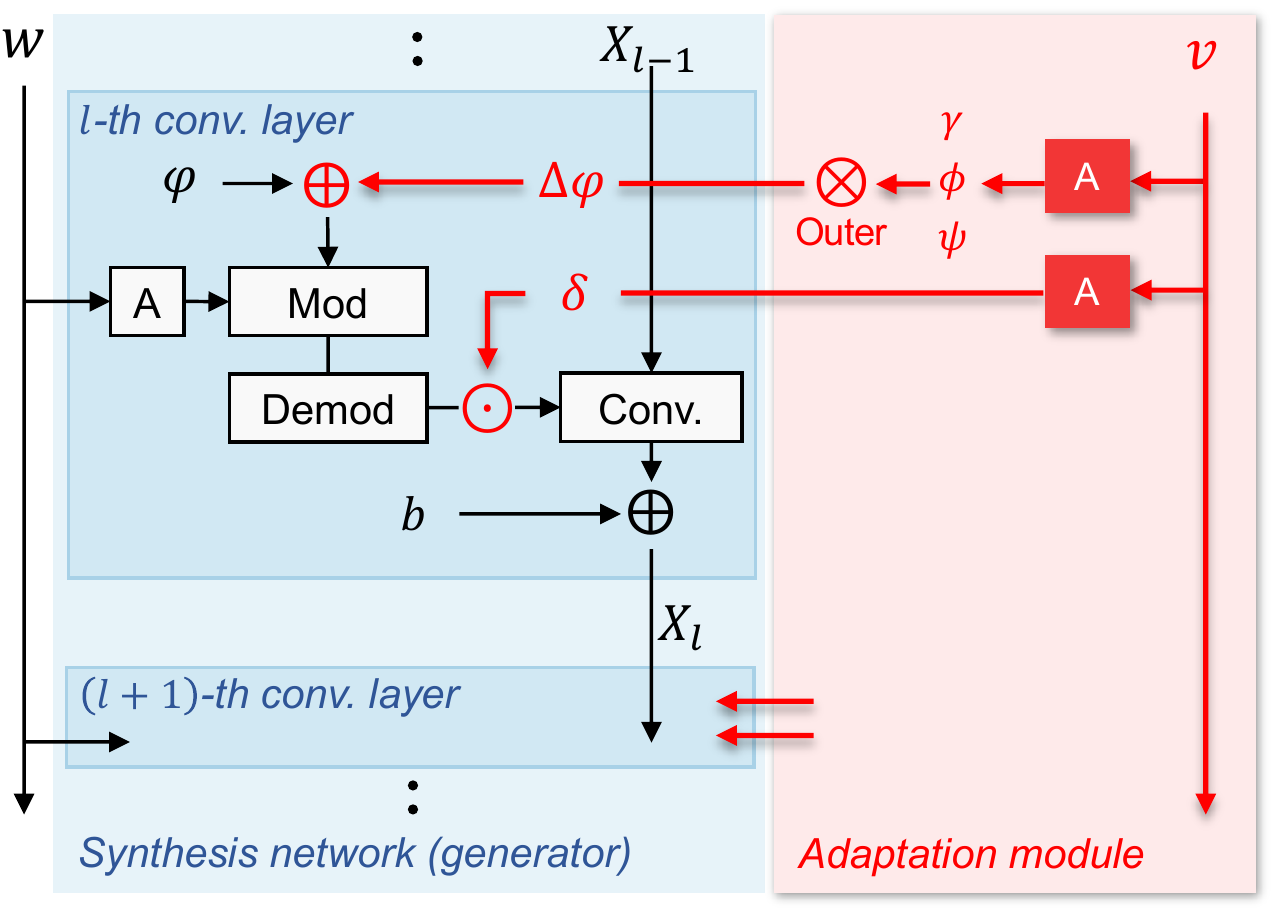}
\vspace{-3mm}
\caption{
Parameter modulation of the generator by the adaptation module. 
}
\label{fig:detail}
\vspace{-5mm}
\end{figure}

\paragraph{Modulation-aware Generator}
For the generator network, we adopt the StyleGAN2 generator architecture, which consists of a mapping network and a synthesis network, with simple architectural changes, which will be described in the next paragraph. 
Here, we first describe the original StyleGAN2 generator.
As illustrated in \Fig{\ref{fig:framework}}, the mapping network $M_G$ converts a latent vector $z$ to an intermediate latent vector $w$, which is fed to the convolutional layers of a synthesis network.
As shown in \Fig{\ref{fig:detail}}, a synthesis network has multiple convolutional layers.
We denote the filter weights of the $l$-th convolutional layer as $\varphi \in \mathbb{R}^{C_\textrm{in}\times C_\textrm{out} \times k}$ where $C_\textrm{in}$ is the input channel size, $C_\textrm{out}$ is the number of output filters, and $k$ is the spatial size of the filter.
In the original StyleGAN2 architecture, the $l$-th convolutional layer transforms an input feature $X_{l-1}$ from the previous layer into an output feature $X_l$ as:
\begin{equation}
    X_l = X_{l-1} *  f\left(\varphi,\textrm{A}\left({w}\right)\right) + {b},
\end{equation} 
where $*$ is the convolution operator and $b$ is the bias of the $l$-th convolutional layer.
$f(\cdot)$ is a composited function defined as $f=\textrm{Demod}\circ\textrm{Mod}(\cdot)$, where $\textrm{Mod}(\cdot)$ and $\textrm{Demod}(\cdot)$ are the modulation and demodulation operations, respectively. $\textrm{A}(\cdot)$ is an affine transformation layer.
For detailed definitions of the operations and layers, we refer the readers to StyleGAN2~\cite{StyleGAN2}.

In order to provide effective modulation on the generator, we change the architecture of the convolutional layer by introducing additional modulation parameters as:
\begin{equation}
    X_l = X_{l-1} *  \left(\delta \cdot f\left(\varphi + \Delta \varphi,\textrm{A}\left(w\right)\right)\right) + {b},    
    \label{eq:modified_generator}
\end{equation} 
where $\delta\in \mathbb{R}^{C_\textrm{out}}$ is a filter-wise scaling factor applied to the output of the function $f$ as $(\delta\cdot \varphi)_{ijk} = \delta_j \varphi_{ijk}$, where $i$, $j$ and $k$ are indices of the input channel, output channel and spatial location, respectively.
$\Delta \varphi$ is a residual weight added to the filter weight.
The two modulation parameters, $\delta$ and $\Delta \varphi$, basically shift and scale the weights of the convolutional layer.
Note that we modulate the convolution weights without touching the other parts of the generator, e.g. the mapping network.
This is aligned with the observations made in recent methods that changing convolutional weights is effective for domain adaptation~\cite{StyleAlign,StyleSpace,toonify}. 
Our approach also uses filter-wise scaling instead of a general linear transformation for $\varphi$, which is motivated by the observation of Wu \textit{et al.}~\cite{StyleSpace} that images can be successfully modified by scaling intermediate feature maps of a generator in a channel-wise manner.
This filter-wise scaling significantly reduces the number of parameters, while still achieving effective domain adaptation. We highlight our modification as red text in \Fig~{\ref{fig:detail}}.

\paragraph{Efficient Adaptation Module}
We then introduce our adaptation module that estimates the modulation parameters $\Delta \varphi$ and $\delta$. 
See \Fig{\ref{fig:detail}} for its network architecture.
Given a target domain one-hot vector $c$, we transform the vector $c$ into a latent vector $\upsilon$ using a multi-layer perceptron (MLP)-based mapping network. 
The latent vector $\upsilon$ is then fed to the affine layers of the adaptation module, resulting in the modulation parameters $\Delta \varphi$ and $\delta$.
Each affine layer is composed of a single fully-connected layer.


Estimating the residual parameters $\Delta \varphi$ for all the layers poses a significant burden in memory and other training resources, as the StyleGAN2 generator has nearly 2.3M parameters per convolutional layer.
To tackle this issue, we represent the residual parameter $\Delta \varphi$ with its rank-1 tensor decomposition~\cite{PEDConv} as 
\begin{equation}
    \label{eq:PEDConv}
    \Delta \varphi = \gamma \otimes \phi \otimes \psi,
\end{equation}
where $\gamma \in \mathbb{R}^{C_\textrm{out}}$, $\phi \in\mathbb{R}^{C_\textrm{in}}$ and $\psi \in\mathbb{R}^k$ are decomposed 1D vectors.
Using rank-1 decomposed vectors significantly reduces the burden of our adaptation network by only estimating three 1D vectors, instead of a high-dimensional tensor $\Delta \varphi$.
Our rank-decomposed representation is more efficient than channel-wise residuals~\cite{HyperStyle}. Refer to the Suppplemental Document for details.

\subsection{Training}
\label{sec:training}


\paragraph{Initialization}
We train our adaptation module to estimate the rank-1 decomposed modulation parameters $\gamma$, $\phi$, $\psi$ and the filter-wise scaling factor $\delta$.
The generator network is fixed during training the adaptation module.
For a warm start of the training process, we initialize our adaptation module so that the modulation parameters do not significantly change the original weights in the pretrained generator network.
Specifically, we initialize the weights of the affine layers of the adaptation module with random initialization attenuated by a linear scalar $0.01$. Furthermore, the bias parameter of the $\delta$-estimation affine layer is set to one to make $\delta$ close to one.
For $\gamma$, $\phi$ and $\psi$, the biases of the corresponding affine layers are initialized to $1$,$1$, and $0$ respectively, resulting in $\Delta \varphi$ close to zero.

\paragraph{Multi-domain Adaptation Loss}
For few-shot multi-domain adaptation, we design our training loss as a weighted sum of three loss functions:
\begin{equation}
    \mathcal{L} = \lambda_\textrm{contra}\mathcal{L}_\textrm{contra} + \lambda_\textrm{MTG}\mathcal{L}_\textrm{MTG} + \lambda_\textrm{ID}\mathcal{L}_\textrm{ID},
\end{equation}
where $\lambda_\textrm{contra}$, $\lambda_\textrm{MTG}$ and $\lambda_\textrm{ID}$ are the balancing weights.
$\mathcal{L}_\textrm{contra}$, $\mathcal{L}_\textrm{MTG}$ and $\mathcal{L}_\textrm{ID}$ are contrastive-adaptation loss, MTG loss and an identity loss, respectively.
In our experiments, we set $\lambda_\textrm{contra}=1, \lambda_\textrm{MTG}=1$ and $\lambda_\textrm{ID}=3$ if both source and target domains contain faces, and $\lambda_\textrm{ID}=0$ otherwise. 

First, our novel contrastive-adaptation loss $\mathcal{L}_\textrm{contra}$ promotes keeping the distinct characteristics of different target domains.
The key idea behind the contrastive-adaptation loss is to make the synthesized image be similar/different with a target training image, for positive/negative pairs, which means that they are in the same/different target domains. 
We denote a training image and a synthesized image at a target domain $c$ as $I_c$ and $\hat{I}_c(w)$, respectively, where $w$ is a StyleGAN2 latent vector. 
We evaluate the similarity between the two images using cosine similarity in a recently-proposed CLIP embedding space~\cite{CLIP}.
Specifically, the contrastive-adaptation loss $\mathcal{L}_\textrm{contra}$ is formulated as  
\begin{equation}
    \mathcal{L}_{\mathrm{contra}} = -\log \frac{\exp (l_{\mathrm{pos}}/\tau)}{\exp (l_{\mathrm{pos}}/\tau) + \sum_{j} \mathds{1}_{[j\neq c]} \exp (l_{\mathrm{neg}}^j/\tau))},
\end{equation}
where $\tau$ is a temperature parameter, which is set to one.
$l_\mathrm{pos}$ and $l_\mathrm{neg}^j$ are similarities of positive and negative pairs defined as:
\begin{eqnarray}
l_\mathrm{pos} &=& \textrm{sim}(E_\mathrm{CLIP}(I_c), E_\mathrm{CLIP}(\hat{I}_c(w)), \nonumber \\
l_\mathrm{neg}^j &=& \textrm{sim}(E_\mathrm{CLIP}(\textrm{Aug}(I_j)), E_\mathrm{CLIP}(\hat{I}_c(w))),
\end{eqnarray}
where $\textrm{sim}(\cdot)$ is cosine similarity and $E_\mathrm{CLIP}$ is a CLIP encoder~\cite{CLIP}. 
$\textrm{Aug}(\cdot)$ is a horizontal-flip and color-jitter augmentation function for stable training~\cite{FuseDream}.

For $\mathcal{L}_\textrm{MTG}$, we adopt the training loss of MTG~\cite{MTG}, which is defined as a combination of a reconstruction loss and CLIP-based losses.
To train our adaptation module, we made a few modifications to the original MTG loss to inject domain dependency, which is detailed in the Supplemental Document. 
The identity loss $\mathcal{L}_\textrm{ID}$ measures the similarity of faces between synthesized and source-domain images in the representation space of a face recognition network~\cite{ArcFace}.
Also, refer to the Supplemental Document for its details.

%% file: e_experiments.tex






\begin{figure*}[!t]
\includegraphics[width=0.9\linewidth]{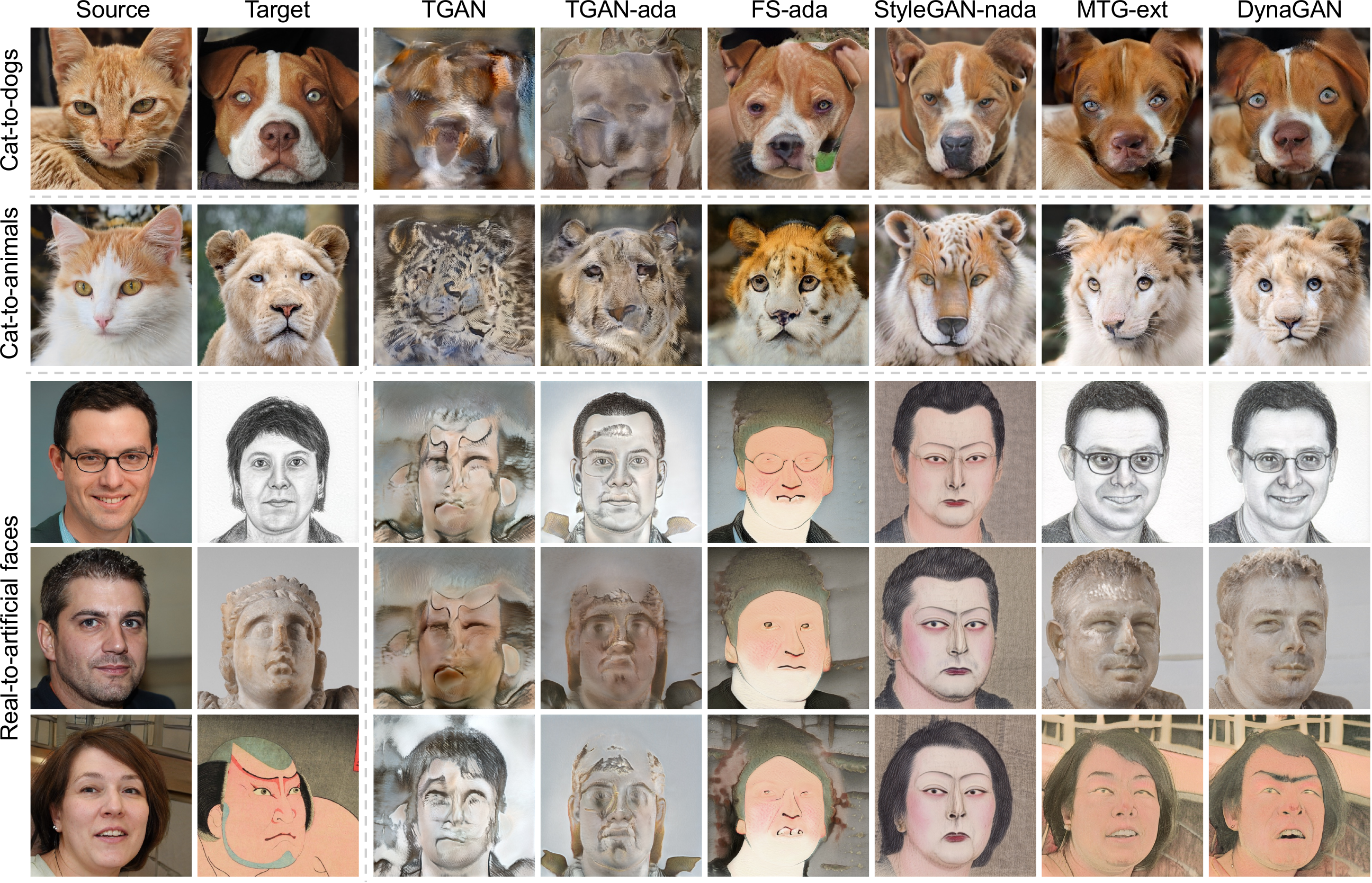}
\vspace{-3mm}
\caption{Qualitative comparison against state-of-the-art few-shot domain adaptation methods: TGAN \cite{TGAN}, TGAN-ada \cite{StyleGAN-ada}, FS-ada \cite{FS-ada}, StyleGAN-nada \cite{StyleGAN-nada}, and MTG-ext \cite{MTG}.
Targets in the 1st, 2nd, 4th and 5th rows: Pixabay (Doz777, PhotoGranary) [Pixabay License],
The Metropolitan Museum of Art [Public Domain], and
ARC Collection, Ritsumeikan University (arcBK01-0042\_37).
}
\label{fig:qual_cmp}
\vspace{-3mm}
\end{figure*}

\section{Experiments}
\label{sec:experiments}
\subsection{Implementation Details}
For training the adaptation module, we use the Adam optimizer \cite{Adam} with coefficients $\beta_1=0.0$ and $\beta_2=0.99$. We set the learning rate to $0.002$ and the batch size to $4$.
When we compute the loss functions, we should embed an image into a CLIP representation space, for which we use `ViT-B/16' and `ViT-B/32' CLIP encoder models~\cite{CLIP} and add their results as done in MTG~\cite{MTG}.

At the inference time, we apply the style mixing of MTG~\cite{MTG} to our generator shown in \Fig{\ref{fig:framework}} in order to better reflect the style of a target domain.
Specifically, for \emph{fine-scale} layers of our synthesis network, we use a latent vector $w_c$  estimated from a target-domain training image 
rather than using a latent vector $w$ from the mapping network of the generator $M_G$
to more faithfully transfer texture styles of the target domain.
For the rest of the layers, we use the latent vector $w$.
More details on the implementation can be found in the Supplemental Document.

\subsection{Comparison}

\paragraph{Comparison Methods}
As \MethodName{} is the first few-shot multi-domain adaptation method, we instead compare our method with five state-of-the-art single-domain adaptation methods: TGAN \cite{TGAN}, TGAN-ada \cite{StyleGAN-ada}, FS-ada \cite{FS-ada}, StyleGAN-nada \cite{StyleGAN-nada}, and MTG-ext, which is our extension of MTG~\cite{MTG}.
While the original MTG is a single-shot adaptation method, we extend it to handle few-shot training images by slightly modifying its loss and to use a particular training image for its style-mixing step.
More details on MTG-ext can be found in the Supplemental Document.

\paragraph{Experimental Datasets}
To evaluate the effectiveness of \MethodName{} compared to the state-of-the-art methods, we curate three challenging few-shot multi-domain datasets: (a) cat-to-dogs, (b) cat-to-animals, and (c) real-to-artificial faces.
For the three datasets, each image corresponds to an \emph{individual} target domain, meaning that each target domain has only one sample image.
Refer to the Supplemental Document for the experiments that use multiple samples per domain.
For (a) the cat-to-dogs dataset, we use five target images sampled from the AFHQ Dog dataset~\cite{StarGANv2} and the source domain is AFHQ Cat dataset~\cite{StarGANv2}.
In (b) the cat-to-animals dataset, we have 10 target images of different animal species from the AFHQ Wild dataset~\cite{StarGANv2}. AFHQ Cat dataset~\cite{StarGANv2} is used again for the source domain.
For (c) the real-to-artificial faces, we use nine target images with diverse forms consisting of three sketch images\footnote{We synthesized sketch images from the Sketch dataset using FS-ada~\cite{FS-ada} instead of directly using the dataset.}, three samples from the MetFace dataset~\cite{StyleGAN-ada}, and three Ukiyo-e images~\cite{ukiyoe}.
We use the FFHQ~\cite{StyleGAN} dataset as the source domain.

\paragraph{Qualitative Assessment}
\Fig{\ref{fig:qual_cmp}} shows synthesized images using each of the adapted models on the three target datasets.
TGAN~\cite{TGAN} and TGAN-ada~\cite{StyleGAN-ada} fail to learn faithful distributions of the target-domain images, resulting in low-fidelity synthesis. 
FS-ada~\cite{FS-ada} and StyleGAN-nada~\cite{StyleGAN-nada} produce images corresponding to average styles of target domains, losing the distinct characteristics of different target domains. 
MTG-ext achieves better results with more faithful textures than the previous methods thanks to the style mixing technique. 
However, it suffers from unnatural global structures, e.g., the deformed dog head in the first row of \Fig{\ref{fig:qual_cmp}}.
\MethodName{} achieves the high-quality images via our adaptation module that dynamically adapts the generator. 

\paragraph{Quantitative Assessment}
We perform quantitative evaluation using Fr\'{e}chet inception distance (FID)~\cite{FID}, inception score (IS)~\cite{IS} and kernel inception distance (KID)~\cite{KID}. 
For the real target-domain images, we use 171 dog images, 483 animal images, and 900 face images in the three datasets. 
We use 50,000 synthesized images for each method.
\Tbl{\ref{table:quantitative}} shows that \MethodName{} achieves superior results to all the other methods, especially for target datasets with large variations:  cat-to-animals dataset and real-to-artifical faces dataset.



\paragraph{Separate Models for Multi-domain Adaptation}
\begin{figure}[!t]
  \includegraphics[width=0.95\linewidth]{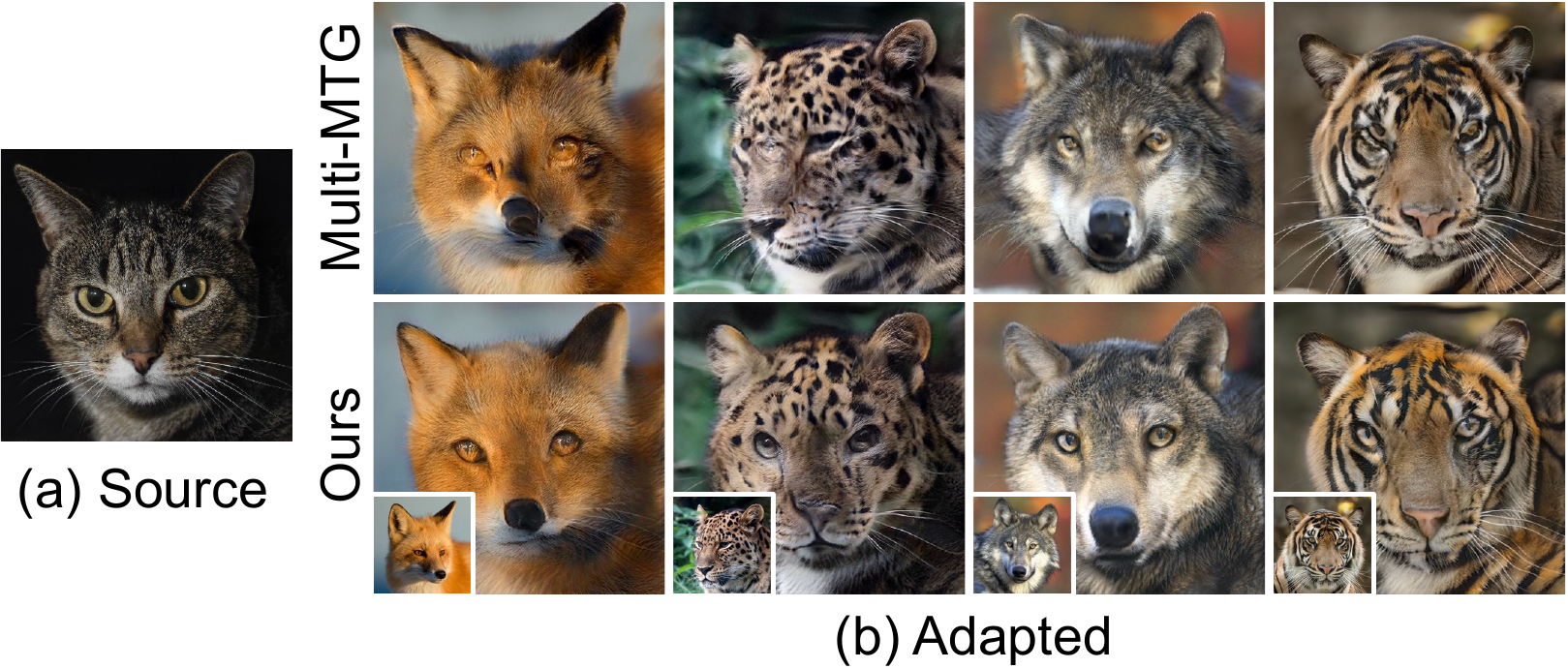}
  \vspace{-3mm}
  \caption{
  Qualitative comparison with multi-MTGs on the cat-to-animals dataset.
  Multi-MTGs consists of multiple MTG~\cite{MTG} models separately trained on each target domain.
  Despite using a single model, our method represents each domain more effectively.
Insets from left to right: Pixabay (Pexels, vinzling, WikiImages, Gregorius\_o) [Pixabay License].
  }
  \label{fig:multiMTG}
\end{figure}

\begin{figure}[!t]
  \includegraphics[width=0.98\linewidth]{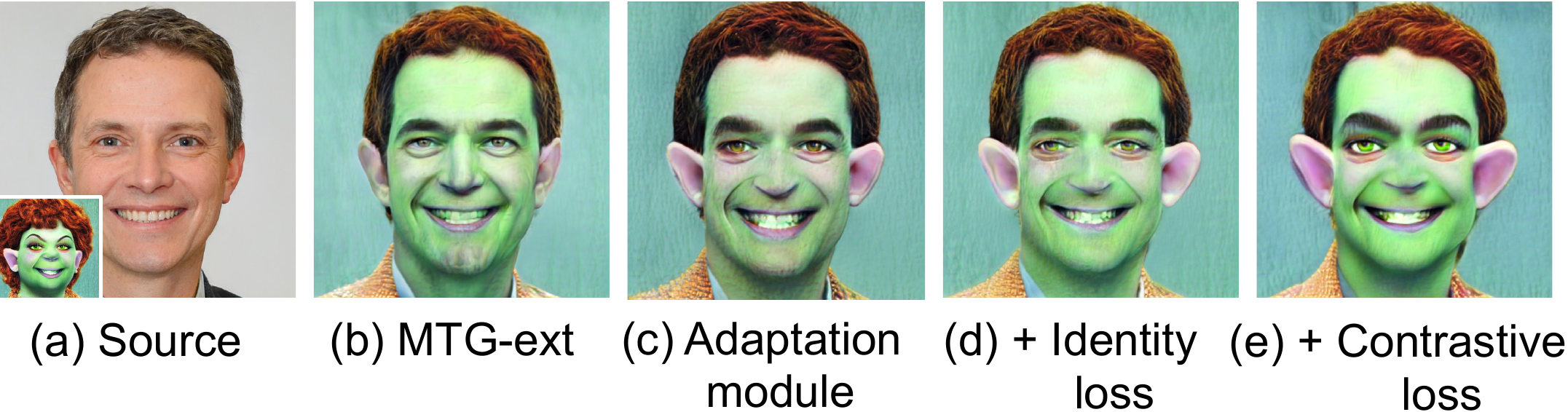}
  \vspace{-3mm}
  \caption{
  For (a) the source image, (b) our baseline model (MTG-ext) fails to synthesize an image that faithfully reflects the style of the target domain.
  (c) Our adaptation module allows us to reflect the style of the target domain by means of dynamic modulation. (d) Identity loss helps preserving the face identity. (e) Our contrastive-adaptation loss further helps reflect domain-specific attributes.
  } 
  \vspace{-3mm}
  \label{fig:ablation}
\end{figure}
As mentioned in \Sec{\ref{sec:introduction}}, a na\"ive approach to multi-domain adaptation is to train a separate model, here we used MTG~\cite{MTG}, for each target domain, called multi-MTGs.
To compare \MethodName{} against the na\"ive approach, we train a separate MTG model for each target domain in the cat-to-animals dataset. 
Note that with 10 different target domains in the cat-to-animals dataset, multi-MTGs result in $10\times$ network parameters compared to the original MTG~\cite{MTG}, resulting in 302.8M parameters. Note that \MethodName{} has only 39.6M parameters, being more light-weight.
\Fig{\ref{fig:multiMTG}} shows a qualitative comparison between \MethodName{} and multi-MTGs.
\MethodName{} produces higher-quality results better reflecting the distinct attributes of the target domains using a single generator and an adaptation module. 


\subsection{Ablation Study}

\Fig{\ref{fig:ablation}} shows an ablation study where we add the components of \MethodName{} one by one to our baseline model, which does not use our adaptation module and the contrastive-adaptation loss.
Here, the target dataset is curated with 5 different images with diverse styles sampled from StyleGAN-nada~\cite{StyleGAN-nada}.
Our adaptation module significantly improves the quality of synthesized images with more plausible shapes of human faces. 
Our contrastive-adaptation loss leads to a result that better reflects domain-specific attributes.


\subsection{Applications}
Here, we demonstrate applications of \MethodName{}: domain interpolation, image-to-image translation, adaptation degree control, and semantic image manipulation.

\begin{figure}[!t]
  \includegraphics[width=0.95\linewidth]{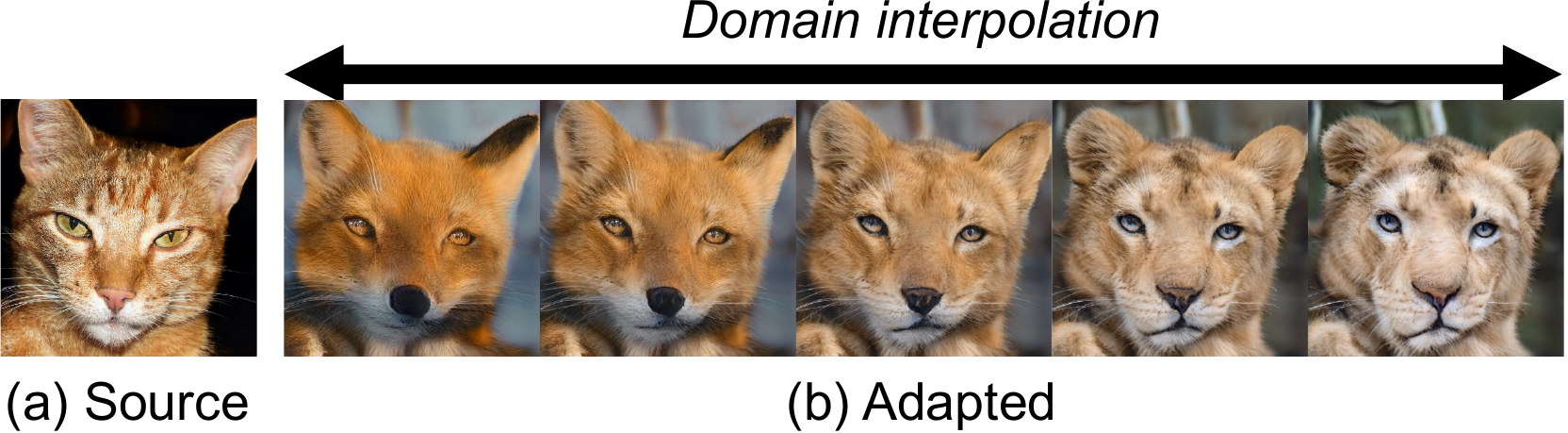}
  \vspace{-3mm}
  \caption{ \MethodName{} supports domain interpolation. 
  For (a) the source image of a cat, we synthesize images with different target domains, here from fox to lion. 
  }
  \vspace{-3mm}
  \label{fig:interpolation}
\end{figure}

\paragraph{Domain Interpolation} 
Our adaptation module provides a smooth transition between two target domains by interpolating the target-domain vector $c$.
\Fig{\ref{fig:interpolation}} shows an example of domain interpolation where the source domain is cat, and we interpolate the target-domain vector from fox to lion.

\begin{figure}[!t]
  \includegraphics[width=0.95\linewidth]{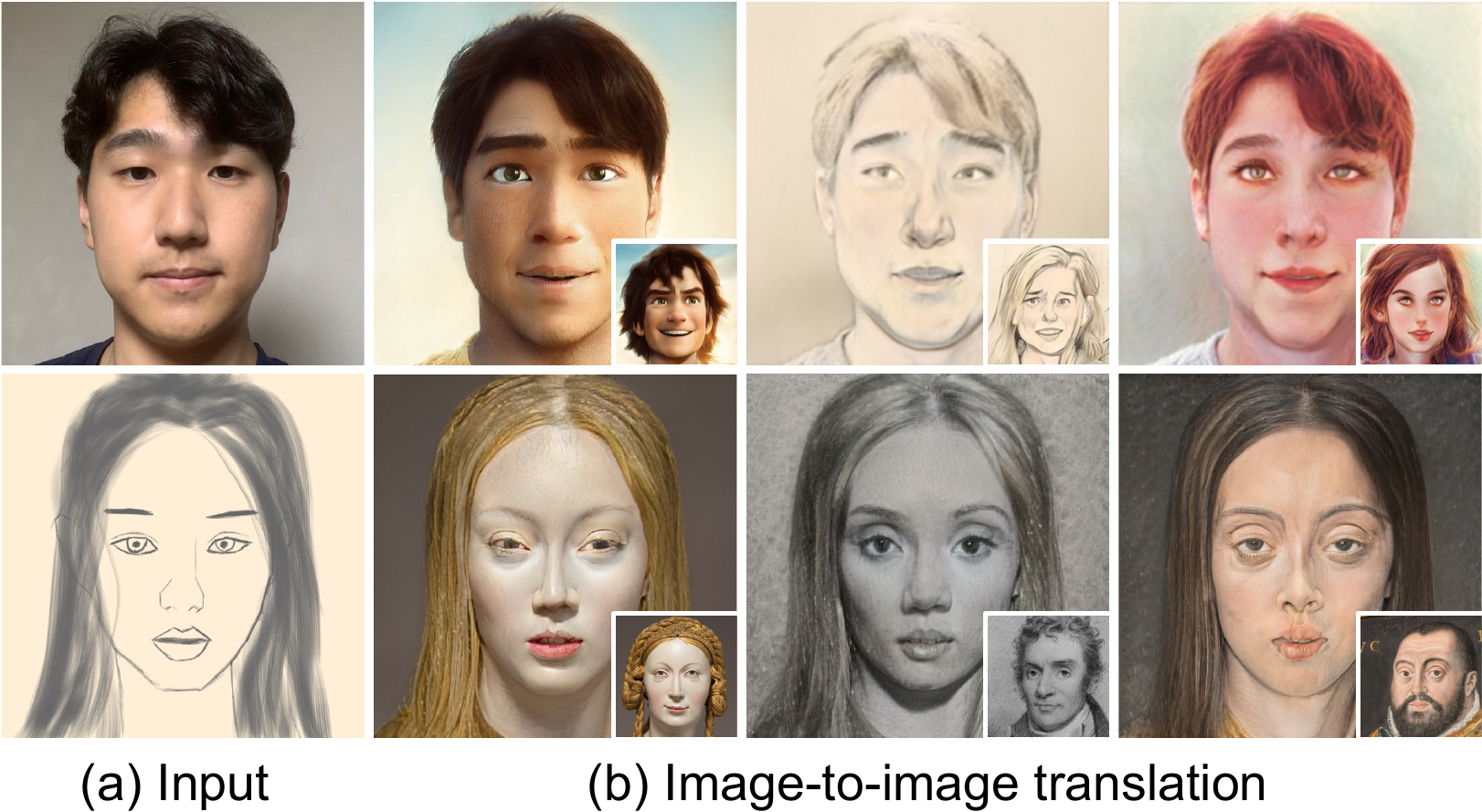}
  \vspace{-0.3cm}
  \caption{Image-to-image translation of real-world images. 
  The inset images in the translation results are training samples of the target domains.
1st, 2nd and 3rd insets in the 2nd row: The Metropolitan Museum of Art [Public Domain].
  }
  \label{fig:transition}
  \vspace{-2mm}
\end{figure}

\paragraph{Image-to-image Translation} 
Image-to-image translation is a task that transfers an image of a source domain to a target domain.
By combining \MethodName{} with GAN inversion, we demonstrate successful image-to-image translation of real-world images. 
\Fig{\ref{fig:transition}} shows results of image-to-image translation.
We invert the input images to GAN latent codes using an off-the-shelf GAN inversion method~\cite{e4e}, and synthesize multi-domain images. 



\begin{figure}[!t]
  \includegraphics[width=0.9\linewidth]{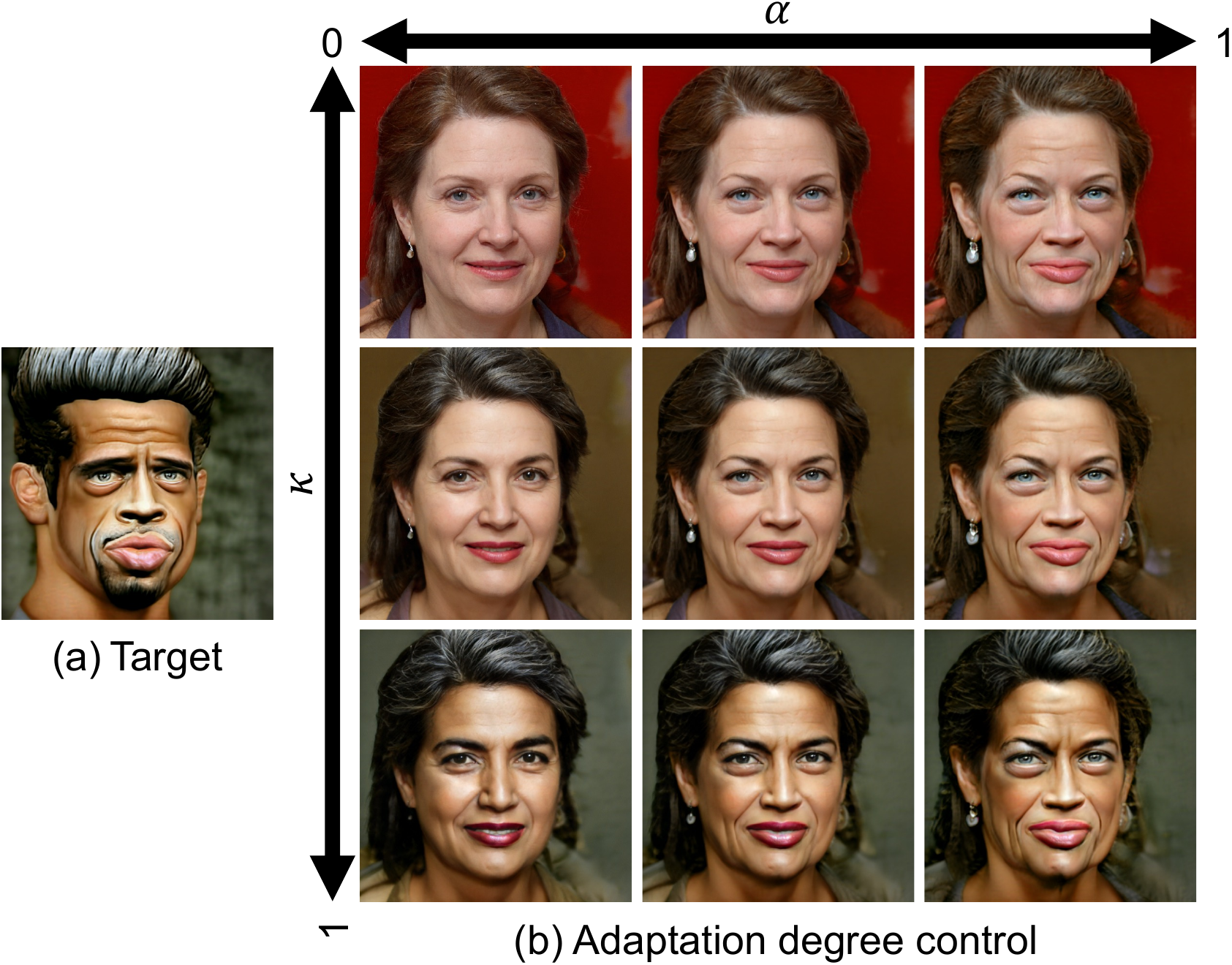}
  \vspace{-3mm}
  \caption{
  Adaptation degree control using the adaptation control parameter $\alpha$ and the style mixing parameter $\kappa$.
  }
  \label{fig:control}
  \vspace{-3mm}
\end{figure}

\paragraph{Controllable Degree of Adaptation} 
\MethodName{} allows to control the degree of adaptation at the inference time by adjusting (1) the modulation parameters $\Delta \varphi$ and $\delta$ in \Eq{\ref{eq:modified_generator}} and (2) the degree of style mixing. 
Specifically, we can scale the modulation parameters as $\Delta \varphi \leftarrow \alpha \Delta \varphi$ and $ \delta \leftarrow \alpha \delta + (1-\alpha)$, where $\alpha$ is the control parameter that adjusts the degree of adaptation.
For the degree of style mixing, we compute an \emph{interpolated} latent vector $\hat{w}$ between $w$ and $w_c$ as $\hat{{w}} = (1-\kappa){w} + \kappa({w}_c)$
where $\kappa$ is the control parameter.
We feed $\hat{w}$ instead of $w_c$ to fine-scale convolutional layers of the generator.
\Fig{\ref{fig:control}} shows results with varying control parameters of $\alpha$ and $\kappa$ from $0$ to $1$.
$\alpha$ controls the adaptation degree of shape and $\kappa$ takes care of texture. 



\begin{figure}[!t]
  \includegraphics[width=0.95\linewidth]{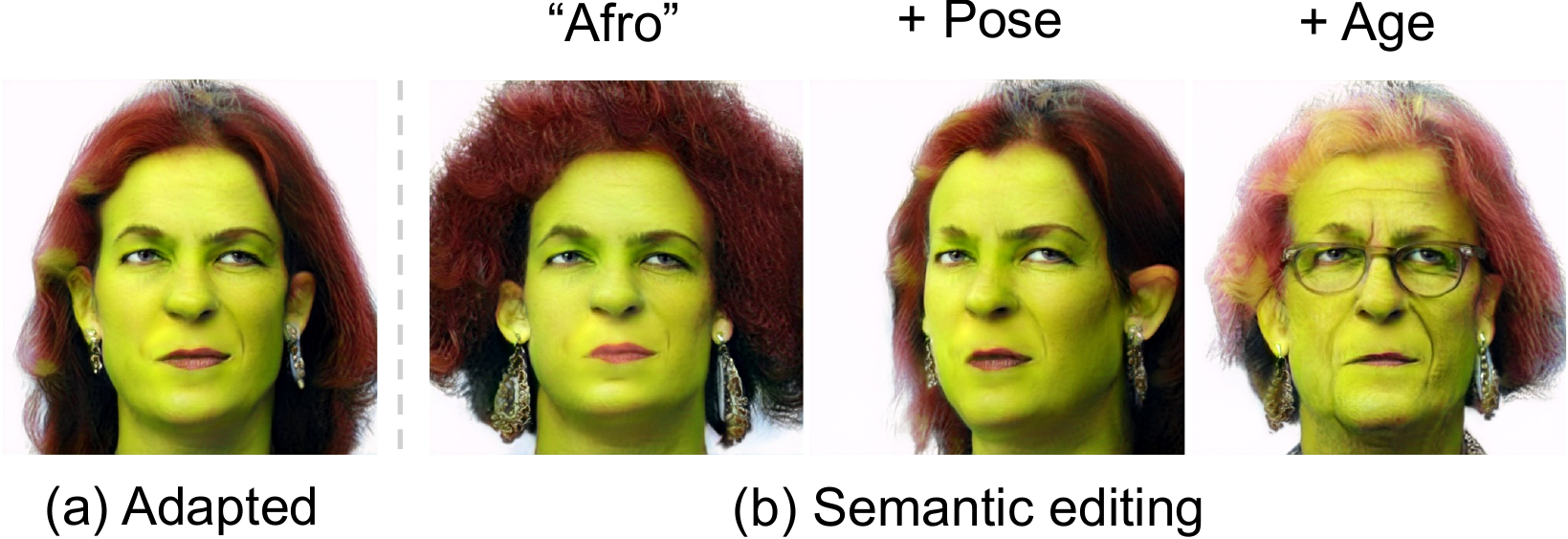}
  \vspace{0mm}
  \caption{
  Semantic editing example. (a) A target-domain image. (b) We add "Afro" hair using the text-driven editing of StyleCLIP~\cite{StyleCLIP}, and change the pose and the age using InterFaceGAN~\cite{InterFaceGAN}.
  }
  \label{fig:edit}
\end{figure}

\paragraph{Semantic Image Manipulation}
As \MethodName{} keeps the mapping network of the generator $M_G$ intact, source and target domains share the common latent space~\cite{StyleAlign}.
This allows manipulation of target-domain images in a GAN latent space in a same way in source-domain images.
\Fig{\ref{fig:edit}} shows an example of semantic image manipulation of a target-domain image where the image is manipulated using StyleCLIP~\cite{StyleCLIP} and InterfaceGAN~\cite{InterFaceGAN}.
\begin{table}%
\caption{Quantitative comparison of different methods on different dataset scenarios. 
}
\label{tab:one}
\begin{minipage}{\columnwidth}
\begin{center}
\scalebox{0.92}{
\begin{tabular}{ccc|c|c|c}
  \hline
                            &                           &               &\multicolumn{3}{c}{Dataset}              \\ \hline            
        \multirow{2}{*}{Model}      &   \multirow{2}{*}{Metric}                  &               &    \multirow{2}{*}{\shortstack{Cat-to-\\ dogs}}   &   \multirow{2}{*}{\shortstack{Cat-to-\\ animals}}                 &      \multirow{2}{*}{\shortstack{Real-to-\\ artificial faces}}                  \\ 
                                            &                           &               &                                               &                                               &          \\ \hline
  \multirow{3}{*}{TGAN}                     & FID                       &$\downarrow$   &   342.97                                      &   179.96                                      &     182.42                  \\ 
                                            & IS                        &$\uparrow$     &   \cellcolor{yellow!60}\textbf{2.40}         &   1.66                                        &     2.03                    \\
                                            & KID$_{\times 10^3}$       &$\downarrow$   &   411.11                                      &   160.90                                      &     106.64                  \\ \hline
  \multirow{3}{*}{TGAN+ada}                 & FID                       &$\downarrow$   &   238.57                                      &   175.46                                      &     178.09                  \\ 
                                            & IS                        &$\uparrow$     &   \cellcolor{yellow!20}\underline{1.61}       &   1.88                                        &     3.11                    \\
                                            & KID$_{\times 10^3}$       &$\downarrow$   &   285.13                                      &   141.79                                      &     118.60                  \\ \hline
  \multirow{3}{*}{FS-ada}                   & FID                       &$\downarrow$   &   73.29                                       &   \cellcolor{yellow!20}\underline{80.80}      &     148.13                  \\ 
                                            & IS                        &$\uparrow$     &   1.31                                        &   1.92                                        &     1.92                    \\
                                            & KID$_{\times 10^3}$       &$\downarrow$   &   52.13                                       &   \cellcolor{yellow!20}\underline{53.72}      &     109.31                  \\ \hline
  \multirow{3}{*}{StyleGAN-nada}            & FID                       &$\downarrow$   &   60.77                                       &   108.50                                      &     142.41                  \\ 
                                            & IS                        &$\uparrow$     &   1.07                                        &   1.18                                        &     1.77                    \\
                                            & KID$_{\times 10^3}$       &$\downarrow$   &   35.61                                       &   84.57                                       &     102.28                  \\ \hline
  \multirow{3}{*}{MTG-ext}                  & FID                       &$\downarrow$   &   \cellcolor{yellow!20}\underline{55.84}      &   89.21                                       &     \cellcolor{yellow!20}\underline{104.59}      \\ 
                                            & IS                        &$\uparrow$     &   1.24                                        &   \cellcolor{yellow!20}\underline{2.37}       &     \cellcolor{yellow!60}\textbf{4.09}        \\
                                            & KID$_{\times 10^3}$       &$\downarrow$   &   \cellcolor{yellow!20}\underline{29.60}      &   58.48                                       &     \cellcolor{yellow!20}\underline{48.77}       \\ \hline
  \multirow{3}{*}{DynaGAN}                  & FID                       &$\downarrow$   &   \cellcolor{yellow!60}\textbf{55.08}        &   \cellcolor{yellow!60}\textbf{38.37}        &     \cellcolor{yellow!60}\textbf{97.23}            \\ 
                                            & IS                        &$\uparrow$     &   1.34                                        &   \cellcolor{yellow!60}\textbf{4.53}         &     \cellcolor{yellow!20}\underline{4.03}           \\
                                            & KID$_{\times 10^3}$       &$\downarrow$   &   \cellcolor{yellow!60}\textbf{23.86}        &   \cellcolor{yellow!60}\textbf{17.36}        &     \cellcolor{yellow!60}\textbf{41.35}          \\ \hline

\end{tabular}}
\label{table:quantitative}
\end{center}

\end{minipage}
\vspace{-0mm}
\end{table}%

%% file: f_conclusion.tex
\section{Conclusion}
\label{sec:conclusion}

In this paper, we propose \MethodName{}, which effectively adapts a pretrained generator to diverse target domains with just a few target images. 
Our adaptation module is light-weight and provides sufficient expressive power for the generator, allowing us to handle multiple target domains. 
Our contrastive loss further encourages generating distinct attributes of different domains. 
We demonstrate that \MethodName{} can effectively handle multiple target domains and achieve superior results to previous state-of-the-art methods.

\begin{figure}[!t]
  \includegraphics[width=0.95\linewidth]{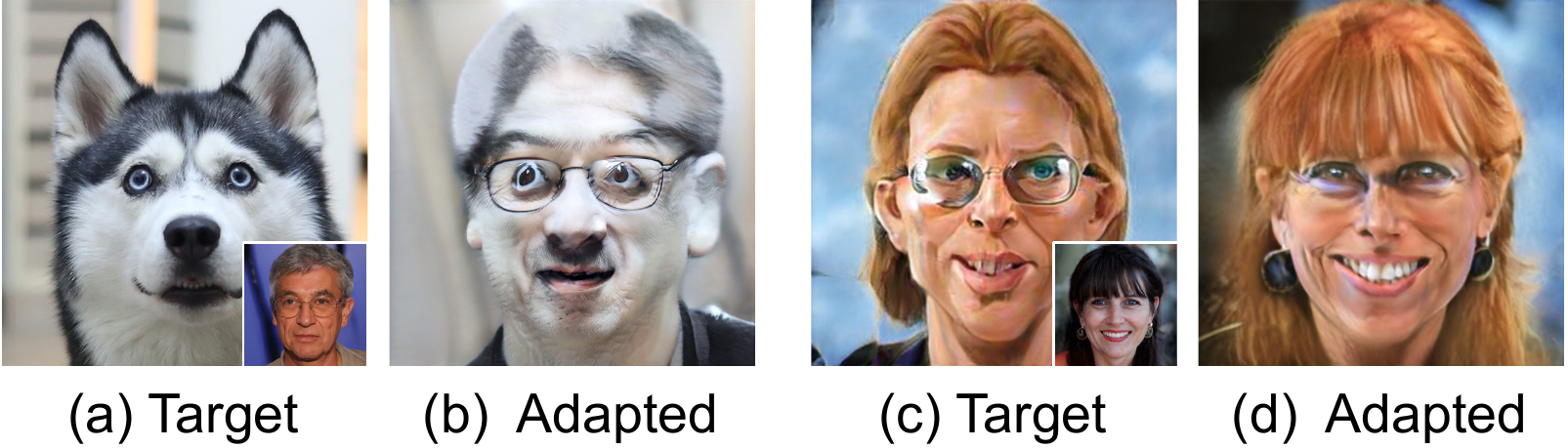}
  \vspace{-3mm}
  \caption{
    Failure cases of \MethodName{}. (a) and (c) are target-domain training images, while (b) and (d) are synthesized results by adapted generators. The inset images in (a) and (c) are source-domain images generated from the same latent codes used for (b) and (d).
Target in (a): Pixabay (Joe007) [Pixabay License].
}
  \label{fig:limitation}
\end{figure}

\paragraph{Limitations}
\MethodName{} fails when there exists a large domain gap between the source and target domains, such as human faces and dogs, as shown in \Fig{\ref{fig:limitation}}(a) and (b).
This is because \MethodName{} fundamentally depends on the source-domain knowledge in the pretrained network for domain adaptation.
Also, using extremely few-shot training images may raise overfitting. For example, in \Fig{\ref{fig:limitation}}(c) and (d), even though its source-domain image has no glasses, the adaptation result has thin structures like glasses around the eyes as the target image has glasses.

\begin{acks}
This research was supported by \grantsponsor{IITP}{IITP}{} grants funded by the Korea government (MSIT) (\grantnum[]{IITP}{2021-0-02068}, \grantnum[]{IITP}{2019-0-01906}), an \grantsponsor{NRF}{NRF}{} grant funded by the the Korea government (MOE) (\grantnum[]{NRF}{2022R1A6A1A03052954}), and \grantsponsor{Pebblous}{Pebblous}{}.
\end{acks}